%% file: main.tex
\documentclass[10pt,twocolumn,letterpaper]{article}

\usepackage{iccv}
\usepackage{times}
\usepackage{epsfig}
\usepackage{graphicx}
\usepackage{amsmath}
\usepackage{amssymb}

\input{yyz}
\input{math_commands}

\usepackage[pagebackref=true,breaklinks=true,colorlinks,citecolor=darkblue,linkcolor=brown,bookmarks=false]{hyperref}
\usepackage[figurename=Fig.,tablename=Tab.]{caption}
\iccvfinalcopy 

\def\iccvPaperID{9394} 

\usepackage[sort,nocompress]{cite}

\ificcvfinal\fi

\begin{document}

\title{Pixel Contrastive-Consistent Semi-Supervised Semantic Segmentation}

\def\inst#1{\unskip$^{#1}$}
\def\email#1{{\small\tt#1}}

\author{
Yuanyi Zhong\inst{1}\thanks{Work partly done during an internship at X, The Moonshot Factory.}\;,
Bodi Yuan\inst{2},
Hong Wu\inst{2},
Zhiqiang Yuan\inst{2},
Jian Peng\inst{1},
Yu-Xiong Wang\inst{1}
\\ \\
\inst{1} University of Illinois at Urbana-Champaign
\hspace{3em}
\inst{2} X, The Moonshot Factory \\
\hspace{2.4em}
\email{\{yuanyiz2, jianpeng, yxw\}@illinois.edu}
\hspace{1.4em}
\email{\{bodiyuan, wuh, zyuan\}@google.com}
}

\maketitle
\ificcvfinal\thispagestyle{empty}\fi

\begin{abstract}
We present a novel semi-supervised semantic segmentation method which jointly achieves two desiderata of segmentation model regularities: the label-space consistency property between image augmentations and the feature-space contrastive property among different pixels. We leverage the pixel-level \ltwo loss and the pixel contrastive loss for the two purposes respectively. To address the computational efficiency issue and the false negative noise issue involved in the pixel contrastive loss, we further introduce and investigate several negative sampling techniques. Extensive experiments demonstrate the state-of-the-art performance of our method (\ours) with the DeepLab-v3+ architecture, in several challenging semi-supervised settings derived from the VOC, Cityscapes, and COCO datasets.
\end{abstract}

\input{sec_intro}
\input{sec_related}
\input{sec_method}
\input{sec_exp}

\section{Conclusion}

We propose a novel semi-supervised semantic segmentation approach based on feature-space contrastive learning and label-space consistency training. We also present the negative sampling techniques to improve the efficiency and effectiveness of pixel contrastive learning. Our approach outperforms existing methods on several semi-supervised segmentation benchmarks, suggesting that pixel contrastive-consistent learning is a promising research direction to improve semi-supervised semantic segmentation.

\paragraph{Acknowledgment:}
This work was supported in part by NSF Grant 2106825.

\input{sec_supp}

{\small
\bibliographystyle{ieee_fullname}
\bibliography{bib}
}

\end{document}

%% file: yyz.tex
 \usepackage{enumitem,amsmath,amssymb,amsthm,commath,dsfont,complexity,mathtools,nicefrac,booktabs,graphicx}

\newcommand{\zyy}[1]{#1}

\def\paragraph#1{\smallskip\noindent\textbf{\boldmath #1\,}}

\usepackage{xcolor}

\definecolor{darkblue}{rgb}{0,0.08,0.45}
\definecolor{seaborn_blue}{HTML}{0072B2}
\definecolor{seaborn_green}{HTML}{009E73}

\def\ours{PC$^2$Seg}
\def\ltwo{$\ell_2$ }  

%% file: math_commands.tex
\usepackage{amsmath,amsfonts,bm}

\def\eqref#1{equation~\ref{#1}}

\def\1{\bm{1}}

\def\vu{{\bm{u}}}
\def\vv{{\bm{v}}}

\def\vy{{\bm{y}}}
\def\vz{{\bm{z}}}

\DeclareMathAlphabet{\mathsfit}{\encodingdefault}{\sfdefault}{m}{sl}
\SetMathAlphabet{\mathsfit}{bold}{\encodingdefault}{\sfdefault}{bx}{n}

\def\gL{{\mathcal{L}}}

\def\sR{{\mathbb{R}}}



%% file: sec_intro.tex
\section{Introduction}

Modern deep learning based solutions \cite{chen2017deeplab,chen2018encoder,sun2019deep} to semantic segmentation typically require large-scale pixel-wise annotated datasets \cite{everingham2010pascal,cordts2016cityscapes,lin2014microsoft} (\ie, the high-data regime). When the deep models are trained with limited labeled data (\ie, the low-data regime), however, their performance drops drastically due to over-fitting. The performance drop in the low-data regime and the high annotation cost associated with dense pixel labeling have motivated the community to study semi-supervised segmentation methods \cite{french2020semi,ouali2020semi,zou2020pseudoseg,ke2020guided,zoph2020rethinking}. In the semi-supervised setting, a model is trained with the help of an additional large-scale unlabeled dataset. A successful semi-supervised method is useful in practice -- only a portion of the production data needs to be densely annotated to deliver a satisfactory model.

\begin{figure}[t]
    \centering
    \includegraphics[width=\linewidth]{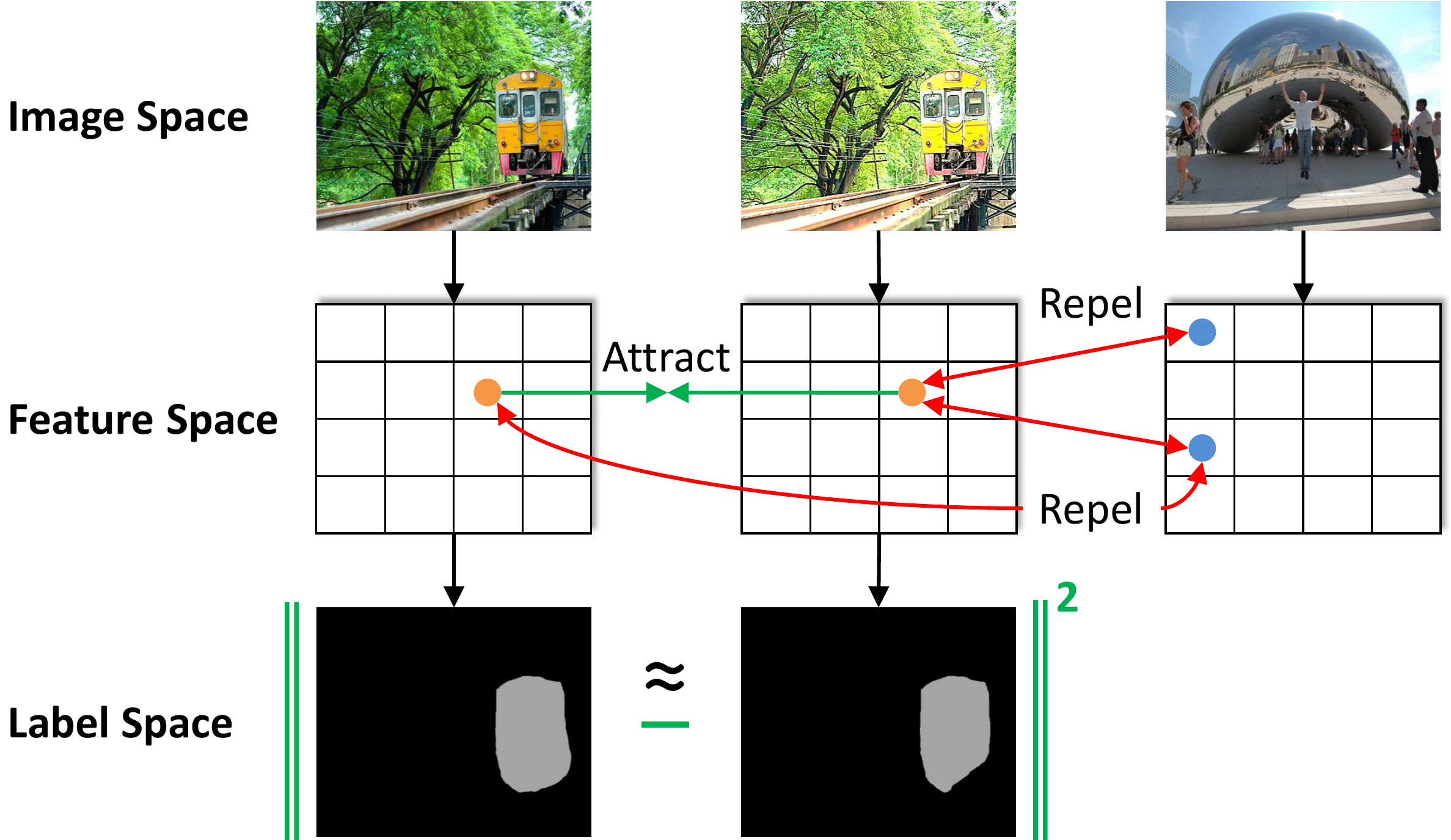}
    \caption{Two desired properties of a semi-supervised segmentation model and our corresponding techniques to simultaneously achieve such properties: (1) Consistency in the label space -- the predicted mask should be invariant to augmentations. We thus leverage the pixel-wise \ltwo loss between two views of the same unlabeled image. (2) Contrastiveness in the feature space -- the features should be able to distinguish similar pixel pairs from dissimilar pairs. To this end, we introduce the pixel contrastive loss that pulls positive pixel pairs closer and pushes negative pairs away.}
    \label{fig:teaser}
    \vspace{-5pt}
\end{figure}

As shown in Fig.~\ref{fig:teaser}, {\em our key insight} is that there are {\em two desired regularity properties} which a semi-supervised segmentation model should possess. The first one is the {\em consistency property in the label space}. The output segmentation mask of the model should be invariant (or equivariant) to the transformations of the input. When an input image goes through two different color augmentations, the segmentation mask should stay the same, since the object semantics and locations are unchanged. The second property is the {\em contrastive property in the feature space}. By ``contrastive,'' we mean that the intermediate features of the model should have the discriminative power to group visually similar pixels together while distinguishing them from visually dissimilar pixels. Such a contrastive property will enable the model to classify pixels into correct semantic categories.

State-of-the-art semi-supervised segmentation methods, in fact, benefit from using the aforementioned label-space consistency property on the unlabeled images, in the forms of invariance under data augmentations \cite{zou2020pseudoseg,french2020semi}, invariance under feature perturbations \cite{ouali2020semi}, consistency of different network branches \cite{ke2020guided}, and model self-consistency \cite{zoph2020rethinking}. However, the explicit application of the feature-space contrastive property and the possibility of the joint label-consistent and feature-contrastive regularization have been largely under-explored.

Based on our insight, this paper explores how to leverage and {\em simultaneously} enforce the consistency property in the label space and the contrastive property in the feature space, leading to a novel {\em pixel contrastive-consistent semi-supervised segmentation} (\ours{}) method. For the label consistency property, we introduce a simple pixel-wise \ltwo consistency loss between the outputs of weakly and strongly augmented views of the same image. For the feature contrastive property, we extend the popular InfoNCE-based contrastive loss \cite{he2020momentum,chen2020simple} and make significant modifications {\em for its use at the pixel level} on an intermediate feature map.

More concretely, the pixel contrastive loss in semantic segmentation faces the unique technical challenges of high computational cost and harmful false negative examples, compared with the standard image-level contrastive learning. The potentially high computation is due to the need of contrasting a large number of pixels. The harm of false negative examples (\ie, when examples of the same class as the anchor are mistakenly chosen as the contrasting examples) has been noted in image-level contrastive learning \cite{chuang2020debiased,huynh2020boosting}. This problem becomes particularly severe in segmentation, where accurate per-pixel predictions are required compared with image classification.
To overcome these issues, we develop simple-yet-effective negative sampling strategies. For each positive pair of pixels, we sample only a moderate number of negative pixels from the mini-batch for efficiency and avoid false negative pixels by a {\em simple cross-image and pseudo-label weighting} heuristic.

Our contributions are three-fold:
\begin{enumerate}[topsep=2pt,itemsep=0pt,parsep=2pt,leftmargin=12pt]
    \item
    We propose the first framework that leverages both pixel-consistency
    (in the label space)
    and pixel-contrastive
    (in the feature space) properties
    for semi-supervised semantic segmentation. We show that these two properties are complementary and their synergy is important.

    \item
    We generalize the existing image-level contrastive learning to pixel-level.
    To overcome the computational cost and false negative difficulties inherent in segmentation tasks, we propose a novel negative sampling technique and investigate its four variants.
    
    \item
    We demonstrate state-of-the-art performance on multiple widely-used benchmarks.
    This is achieved relatively easily by leveraging the proposed framework
    together with standard loss functions and data augmentations,
    without sacrificing efficiency compared to other semi-supervised methods.

\end{enumerate}

%% file: sec_related.tex
\section{Related Work}
\label{sec:related}
\paragraph{Contrastive Learning.}
Self-supervised or unsupervised visual representation learning has been gaining momentum \cite{he2020momentum,chen2020exploring,chen2020simple,chen2020big,grill2020bootstrap}. At the center of the recent progress is the contrastive learning based pretext tasks \cite{he2020momentum,chen2020simple,chen2020big}, which learn representations that discriminate similar image pairs (constructed from different augmentations of the same images) from dissimilar, negative image pairs. A variety of strategies have been investigated to choose appropriate negative pairs. In MoCo \cite{he2020momentum,chen2020improved}, a memory buffer and a momentum encoder are maintained to provide negative samples. In SimCLR \cite{chen2020simple,chen2020big}, the negative samples are the large training mini-batches. Improper negative pairs might hurt learning performance: Particularly related to our negative pixel sampling strategies, \cite{chuang2020debiased,huynh2020boosting,wu2020conditional} have proposed ways to alleviate the bias issue caused by incorrect (false) negative images by modifying the contrastive loss function. \ours{} does not change the loss function but rather samples the negative examples strategically.
SimSiam \cite{chen2020exploring} and BYOL \cite{grill2020bootstrap} find that simple consistency training without negatives can also be effective with carefully designed architectures and training procedures. However, our ablation study shows that \ours{} still benefits from negative samples compared with feature-space consistency training alone.

Pixel-level contrastive learning has not been well-explored until very recently.
\cite{pinheiro2020unsupervised,wang2020dense,xie2020propagate,van2021unsupervised} explore dense pixel-level self-supervised pretraining. They show that pixel-level pretext tasks can transfer better to segmentation than image-level self-supervised learning.
In this paper, we attempt to benefit from pixel-level contrastive learning in the semi-supervised rather than the unsupervised setting. We present a strong semi-supervised approach that jointly optimizes a contrastive loss on the intermediate features and a consistent loss on the output masks.

Contrastive learning can be used in the supervised setting \cite{khosla2020supervised} to improve generalization.
Researchers have attempted to apply supervised contrastive learning for semantic segmentation \cite{wang2021exploring,zhao2020contrastive}. \cite{wang2021exploring} directly leverages pixel contrasting in supervised segmentation training, while \cite{zhao2020contrastive} adopts it during the first supervised stage in a multi-stage semi-supervised setting. Compared with these attempts, we focus on the single-stage semi-supervised setting, where \emph{self-supervised} contrastive loss is jointly applied on the \emph{unlabeled} images branch without using any ground-truth labels.

\paragraph{Semi-Supervised Learning.}
Semi-supervised learning aims at leveraging labeled data and a large amount of unlabeled data. The idea of consistency regularization is inherent in many successful semi-supervised approaches. Iterative pseudo-labeling and re-training \cite{lee2013pseudo,tarvainen2017mean,zoph2020rethinking} implicitly enforces consistent predictions between the current model (student) and its past versions (teacher), leading to smaller entropy and larger class separation of the predicted label distribution.
FixMatch \cite{sohn2020fixmatch}, VAT \cite{miyato2018virtual}, and UDA \cite{xie2020unsupervised} leverage consistency regularization between weak and strong (or local adversarial \cite{miyato2018virtual}) augmented views of the same unlabeled image in a single-stage training pipeline. S4L \cite{zhai2019s4l} explores a self-supervised auxiliary task (\eg., predicting rotations) on unlabeled images jointly with a supervised task.

\paragraph{Semantic Segmentation.}
Semantic segmentation is the task of predicting pixel-level category labels from images. High segmentation accuracy can be reached by deep fully convolutional neural networks (FCNs) trained on large datasets \cite{chen2017deeplab,chen2018encoder,sun2019deep}. In these models, the convolutional nature of FCNs is exploited to generate dense prediction masks.
Following \cite{zou2020pseudoseg,french2020semi}, our work builds upon the widely-used DeepLab-v3+ segmentation model \cite{chen2018encoder}.

\paragraph{Semi-Supervised Semantic Segmentation.}
It is compelling to study semi-supervised segmentation approaches to reduce the mask labeling cost.
\cite{souly2017semi} synthesizes additional training data with generative adversarial networks (GANs). \cite{hung2019adversarial,mittal2019semi} couple an adversarial loss on the predicted masks with the standard supervised loss. In the line of consistency training, the output predictions of the unlabeled images are encouraged to be consistent across different augmented views \cite{french2020semi,kim2020structured,olsson2021classmix,zou2020pseudoseg}, feature embedding perturbations \cite{ouali2020semi}, and different networks as in co-training \cite{ke2020guided,wang2020self}. When the consistency loss is set up explicitly as using one branch's prediction as the target to train the other branch, the former predictions are often called pseudo labels \cite{french2020semi,zou2020pseudoseg}. Self-training, which retrains the model with pseudo labels on the unlabeled data given by the previous iteration of the model, has also shown promising results for semantic segmentation \cite{chen2020naive,zoph2020rethinking,zhu2020improving}. While consistency training is extensively studied, the use of (pixel) contrastive leaning has largely eluded discussion in the semi-supervised segmentation setting.

Finally, alternative ways to reduce the annotation cost includes leveraging various forms of weak labels such as image labels \cite{ahn2018learning,luo2020semi,sun2020mining,zou2020pseudoseg,wang2020self}, bounding boxes \cite{dai2015boxsup}, and scribbles \cite{lin2016scribblesup}, or even purely unlabeled data as in unsupervised segmentation by clustering \cite{ji2019invariant} and self-supervised tasks \cite{zhan2018mix}. We refer interested readers to these works as the problem settings are different.

%% file: sec_method.tex
\begin{figure}[t]
    \centering
    \includegraphics[width=\linewidth]{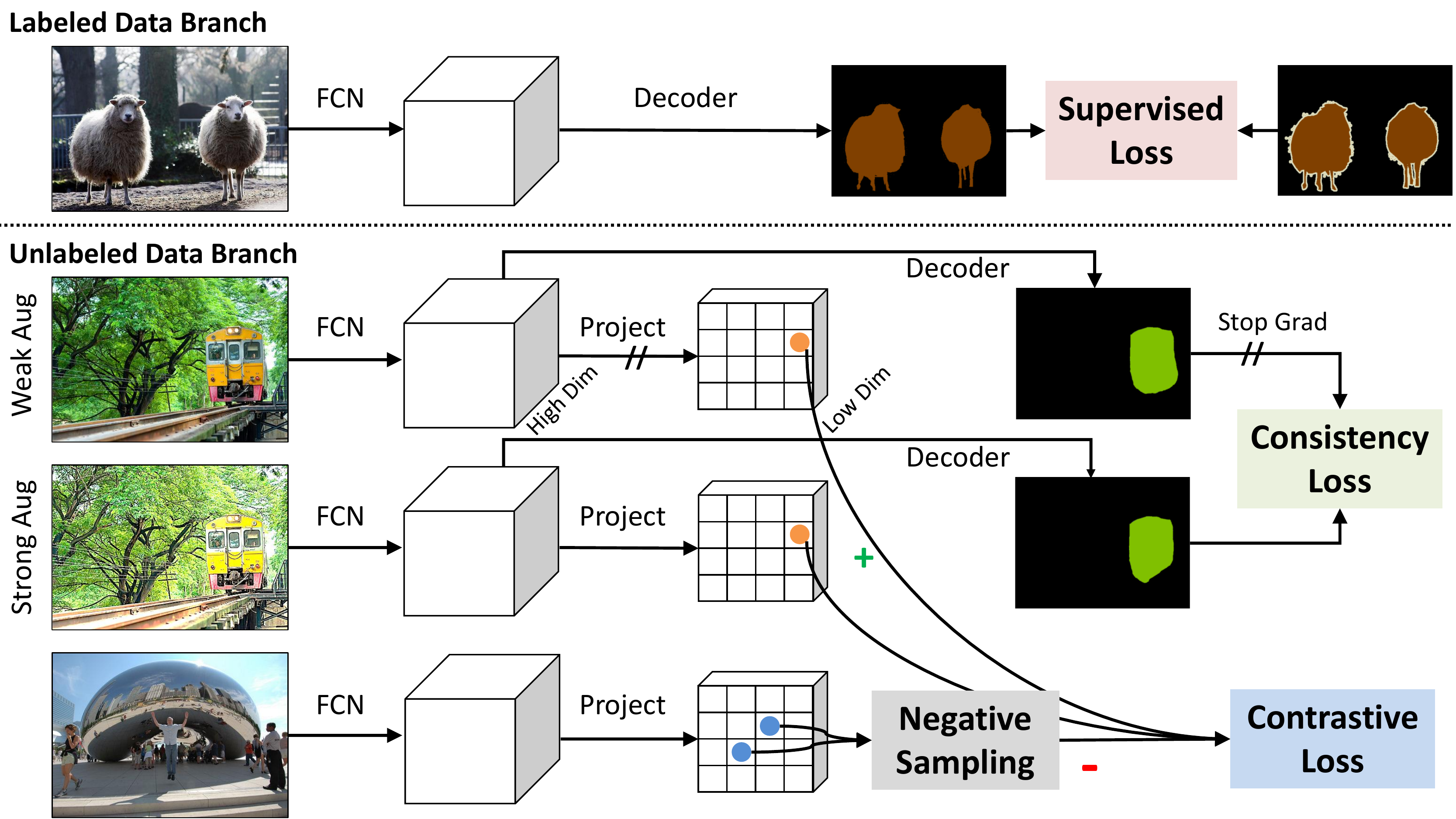}
    \caption{\ours{} overview. The training pipeline has two input streams: One for labeled images and the other for unlabeled images. The images pass through a \zyy{shared} fully convolutional network (FCN) to generate mid-level features and a \zyy{shared} decoder to generate the mask predictions. The labeled branch is trained as usual with the supervised cross-entropy loss. Two augmented views (weak and strong) are generated from unlabeled images. Then we \zyy{use the predicted masks of the weak view as pseudo labels and} impose a consistency loss between the predicted masks of the two views. A pixel contrastive loss is applied on the projected low-dimensional features of an intermediate layer with negative pixels sampled by the strategies in Sec.~\ref{sec:method/neg}.}
    \label{fig:overview}
    \vspace{-10pt}
\end{figure}

\section{Method}
\label{sec:method}

\subsection{Overview}

Fig.~\ref{fig:overview} summarizes the overall architecture and training procedure of our pixel contrastive-consistent semi-supervised segmentation (\ours{}) approach. The entire pipeline consists of two data streams, with one \zyy{stream} for labeled images and the other for unlabeled images. The complete loss function is the sum of the labeled and unlabeled stream losses:
\begin{equation}
    \label{eq:all}
    \gL = \gL^\mathrm{label}(x, y) + \gL^\mathrm{unlabel}(x)
    ,
\end{equation}
where we denote an image as $x$, and its ground-truth segmentation mask as $y$.

For the labeled images, the typical supervised cross-entropy loss between the predictions and the segmentation masks is applied at per-pixel locations. For the unlabeled images, in line with existing semi-supervised \cite{sohn2020fixmatch,zou2020pseudoseg} and self-supervised \cite{chen2020simple,he2020momentum,chen2020exploring,huynh2020boosting} methods, we build a siamese network architecture \zyy{with shared weights} that operates on two augmented views (weak and strong) of a single unlabeled image.
The weak augmentations consist of the usual crop, flip, and resize transformations, while the strong augmentations add color (brightness, contrast, and hue) and Cutout \cite{devries2017improved} transformations on top of the weak augmentations following \cite{zou2020pseudoseg}.
\zyy{We then jointly apply the label consistency loss $\ell^\mathrm{cy}$ on the output masks and the unlabeled pixel constrastive loss $\ell^\mathrm{ce}$ on the intermediate features:}
\begin{equation}
    \label{eq:unlabeled}
    \gL^\mathrm{unlabel} = \sum_i \lambda_1 \ell^\mathrm{ce}(\vz_i, \vz_i^+, \{\vz_{in}^-\}_{n=1}^N) + \lambda_2 \ell^\mathrm{cy}(\hat{\vy}_i^{}, \hat{\vy}_i^+)
    .
\end{equation}
Here we use $i$ to index the pixels, use $\vz_i$, $\vz_i^+$, and $\vz_{in}^-$ to represent the anchor pixel from one augmented view, the positive pixel from the other augmented view, and the negative pixels, respectively. $\hat{\vy}_i^{}$ and $\hat{\vy}_i^+$ are the \zyy{predicted pixel class softmax probabilities} of the two views. $\lambda_1$ and $\lambda_2$ are trade-off hyper-parameters. Exactly which intermediate feature map to apply $\ell^\mathrm{ce}$ is a design choice.

\paragraph{Consistency Loss.} We adopt the normalized \ltwo pixel consistency loss (equivalently, $1 - \text{cosine similarity}$) between the corresponding predicted softmax probabilities of the weak and strong views:
\begin{equation}
\label{eq:consist}
\ell^\mathrm{cy}_i = 
1 - \cos(\hat{\vy}_i^{}, \hat{\vy}_i^+).
\end{equation}
The cosine similarity is $ \cos(\vu, \vv) = \frac{ \vu^\top \vv }{ \norm{\vu}_2 \norm{\vv}_2 } $.
In practice, we sharpen the softmax for the weak branch output $\hat{\vy}_i^{}$ by dividing the logits by temperature $0.5$.
We also put a stop-gradient operation on the weak output as in \cite{chen2020simple,grill2020bootstrap,chen2020exploring,zou2020pseudoseg}, so that the gradients only back-propagate into the strong branch but not the weak branch. 
This effectively makes the weak branch predictions $\hat{\vy}_i^{}$ act as the pseudo labels to train the strong branch.
Empirically we found that our \ltwo consistency loss outperforms the cross-entropy loss in \cite{zou2020pseudoseg}.

\subsection{Pixel Contrastive Loss}

Suppose the feature map, of either the weak branch or the strong branch, has shape $B \times H\times W \times D$, where $B$ is the batch size, $H$ and $W$ are the height and width, and $D$ is the feature dimension. We denote $i\in \{1,2,\ldots,B\times H\times W\}$ as the index of one pixel location on this feature map.
Given the feature vector $\vz_i \in \sR^D $ of the anchor pixel, the goal of contrastive learning is to increase its similarity to a positive pixel $\vz_i^+$ and reduce its similarity to $N$ negative pixels $\{\vz_{in}^-\}_{n=1}^N$. A popular choice of the loss function to achieve such goal is the InfoNCE \cite{hjelm2018learning} loss, which is relevant to the noise contrastive mutual information estimator. When combining the InfoNCE loss with the cosine similarity, we arrive at the pixel contrastive loss in the form of
\begin{equation}
    \label{eq:contrast}
    \ell^\mathrm{ce}_i = -
    \log \frac{ e^{\cos(\vz_i^{}, \vz_i^+) / \tau} }
        { e^{\cos(\vz_i^{}, \vz_i^+) / \tau} + \sum\limits_{n=1}^N e^{\cos(\vz_i^{}, \vz_{in}^-) / \tau} }
    .
\end{equation}
Here $\tau$ is a temperature hyper-parameter to control the scale of terms inside exponential, which is set to $0.07$ throughout our experiments following prior work \cite{chen2020simple,chen2020big}. The use of cosine similarity is shown to be effective in existing contrastive learning work \cite{chen2020simple,he2020momentum,tian2019contrastive,wang2021exploring,grill2020bootstrap,chen2020simple}.

Directly optimizing Eq.~\ref{eq:contrast}, however, is challenging. The first difficulty is that the raw feature vectors can be quite high-dimensional, \eg, the last ResNet backbone feature map has shape $33 \times 33 \times 2,048$ in DeepLab \cite{chen2018encoder}, leading to high memory and computational cost. We therefore utilize a linear projection layer to reduce the feature dimension from $2,048$ to $128$, consistent with \cite{huynh2020boosting,wang2021exploring}. The projection head parameters of the weak and strong branches can be separated or tied together. We found that separated projection heads work slightly better in our experiments. In the weak branch, similar to the consistency loss, a stop gradient operation is inserted before the projection.

The next major difficulty lies in deciding the positive and negative pairs. In the ideal supervised learning setting, \ie, when the ground-truth pixel labels are known, we can simply select the positive pixel from the pixels of the true category of the anchor pixel, and choose the negative pixels as the pixels from different categories \cite{khosla2020supervised}. By contrast, the issue becomes complicated when the labels are unknown as in our setting. Existing work on image-level contrastive learning circumvents the problem by forcing the positive example to come from another augmented view of the same image, while using both views of a large mini-batch \cite{chen2020simple} or a memory bank \cite{he2020momentum} as the source of negative examples.

However, these techniques are not \zyy{easily} generalizable to our task. For contrastive learning at pixel level, we still choose the positive pixel as the corresponding pixel under a random color augmentation. For example, if $\vz_i$ belongs to the weak view, $\vz_i^+$ can be the corresponding pixel in the strong view.
As for the negative pixels, two issues occur if we directly borrow existing approaches in image-level contrastive learning:
(1) Due to the huge number of pixels, we cannot afford to use all the pixels in the mini-batch as negative examples, because the memory and computational cost scale with $O(\text{num\_pixels}^2)$;
(2) Since the task is at pixel-level, the segmentation model can be sensitive to noise from incorrect negative pixels, if we adopt a simple mini-batch or memory bank approach. A pixel of the same category as the anchor pixel might be wrongly chosen as a negative pixel, leading to ineffective or even misleading learning signal.
To resolve these issues, we propose to sub-sample a fixed number of negative pixels for each anchor pixel, and develop several effective sub-sampling strategies as below.

\subsubsection{Negative Sampling Strategies}
\label{sec:method/neg}

Consider the $i$-th anchor pixel. Assume the $N$ negative pixels $\vz_{in}^-$ are sampled without replacement according to a discrete distribution $p_{ij}$, which is defined on a total of $M = B\times H\times W$ candidate pixels. More formally,
\begin{equation}
    \vz_{in}^- \sim \mathrm{Discrete}\left(
      \{\vz_j\}_{j=1}^M; \{p_{ij}\}_{j=1}^M \right)
    .
\end{equation}

Different sampling strategies essentially define different sampling distributions $\{p_{ij}\}_{j=1}^M$. We investigate and compare four of them in this paper, with illustration in Fig.~\ref{fig:samp}:

\begin{figure}
    \centering
    \includegraphics[width=0.95\linewidth]{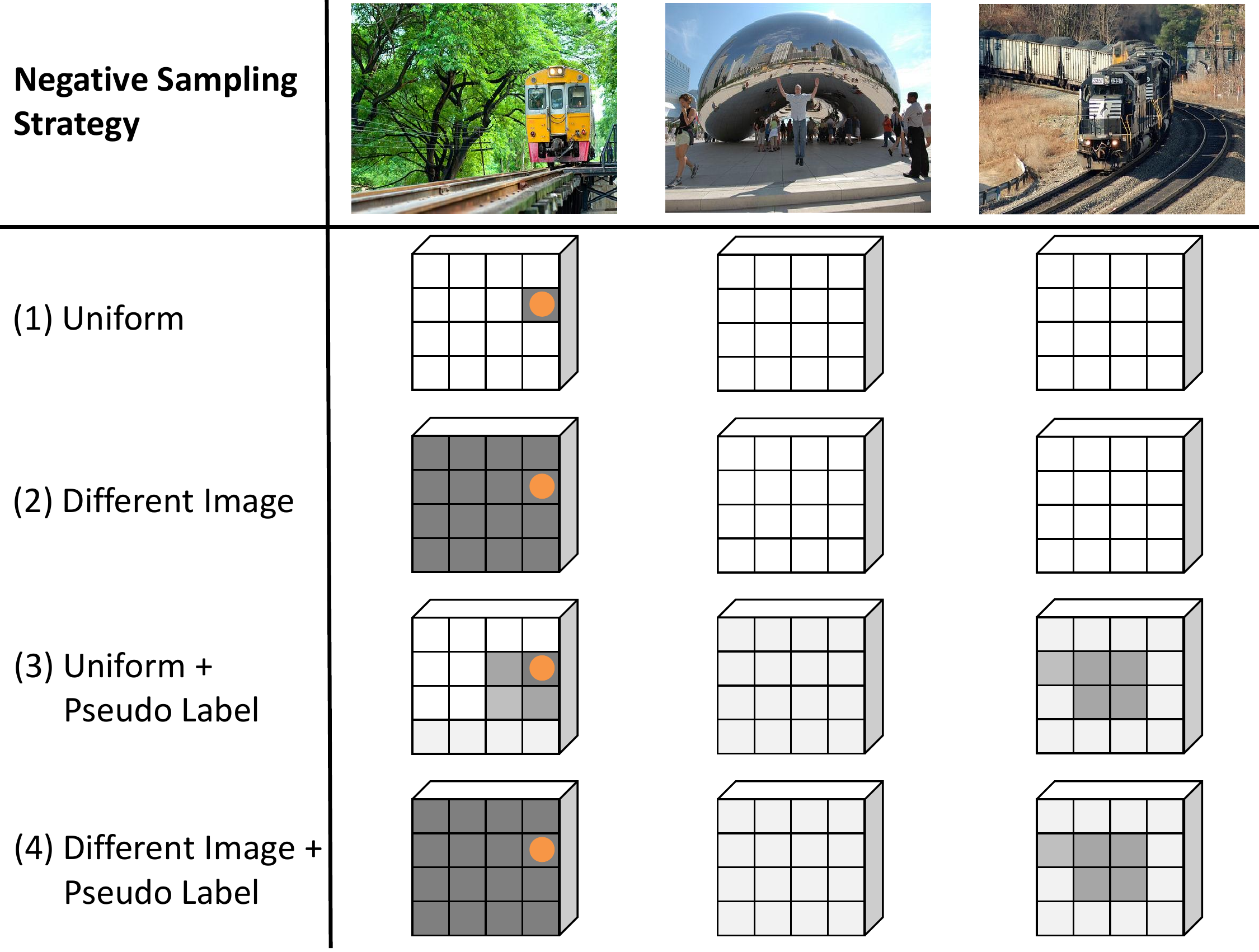}
    \caption{Illustration of negative sampling strategies. The anchor pixel (at the location of the train) is marked with the orange dot. Darker shade means lower chance of sampling that pixel. (1) Uniform strategy only avoids sampling the anchor pixel. (2) Different Image avoids sampling from the same image. In (3)(4), the Pseudo Label correction reduces the chance of sampling the train pixels in the last column.}
    \label{fig:samp}
\end{figure}

\paragraph{Uniform.}
    The most straightforward way is to sample negative pixels uniformly from the mini-batch. Suppose there are $M$ valid pixels in the current mini-batch; each pixel $j=1,2,\ldots, M$ gets a uniformly distributed density,
    $
        p_{ij} = \frac{1}{M}.
    $
    
    As one can imagine, since many pixels on the same image may belong to the same category, the uniform strategy can produce a large number of false negative pixels, which may hurt performance. This issue motivates us to study alternative sampling distribution.

\paragraph{Different Image.}
    One idea to fix the false negative issue is to force the negative pixels to come from a different image, such that the chance of sampling an incorrect negative is reduced. Formally, we denote $I_i$ and $I_j$ as the image IDs of the anchor pixel $i$ and a candidate negative pixel $j$, and $\1\{\cdot\}$ as the indicator function. Pixel $j$ gets a non-zero density only if it belongs to a different image from pixel $i$.
    \begin{equation}
        p_{ij} = \frac{ \1\{I_i \ne I_j\} } { \sum_{k=1}^M \1\{I_i \ne I_k\} }
        .
    \end{equation}

\paragraph{Pseudo-Label Debiased.}
    Another idea to fix the issue is to leverage the model predictions to filter out possible pixels of the same class. This strategy is similar in spirit to \cite{chuang2020debiased,huynh2020boosting}. Suppose $\hat{y}_i$ and $\hat{y}_j$ ($\in \{1,2,\ldots,\text{num\_classes}\}$) are the predicted pseudo labels of pixels $i$ and $j$, respectively (through a bilinear image resizing if necessary). The probability of sampling pixel $j$ as negative pixel can be weighted by $ P(\hat{y}_i \ne \hat{y}_j) $, which biases the sampling process towards pixels that the model believes are from different categories. After a simple derivation, $ P(\hat{y}_i \ne \hat{y}_j) $ can be rewritten as one minus the dot product of the predicted label probability vectors $\vec{\vy}_i$ and $\vec{\vy}_j$.
    \begin{equation}
        p_{ij} = \frac{ P(\hat{y}_i \ne \hat{y}_j) }
        { \sum_{k=1}^M P(\hat{y}_i \ne \hat{y}_k) }
        =
        \frac{ 1 - \vec{\vy}_i^\top \vec{\vy}_j }
        { \sum_{k=1}^M 1 - \vec{\vy}_i^\top \vec{\vy}_k }
        .
    \end{equation}
    
    The derivation is quite straight-forward. Ideally, we want to sample from the discrete measure: $\1\{j: y_i \ne y_j\} = P(y_i \ne y_j) \in \{0, 1\}$. Since the ground-truth is unknown, we may approximate $P(y_i \ne y_j)$ with the pseudo-labels $\hat{y}_i$ and $\hat{y}_j \in \{1,2,\ldots,\text{num\_classes}\}$. Let the predicted probability of pixel $i$ belonging to category $c$ be $ P(\hat{y}_i = c)$ and the vector $\vec{\vy}_i = [P(\hat{y}_i = 1), P(\hat{y}_i = 2), \ldots]$, and then we have
    $
    P(y_i \ne y_j)
        \approx P(\hat{y}_i \ne \hat{y}_j)
        = 1 - P(\hat{y}_i = \hat{y}_j)
        = 1 - \sum_c P(\hat{y}_i = c, \hat{y}_j = c)
        = 1 - \sum_c P(\hat{y}_i = c) P(\hat{y}_j = c)
        = 1 - \vec{\vy}_i^\top \vec{\vy}_j
    .
    $

\paragraph{Different Image + Pseudo-Label Debiased.}
    The above two strategies can be combined to further reduce the chance of sampling false negative pixels. In another word, we ensure that the negative pixels are coming from different images and at the same time reweight them by the pseudo labels.
    \begin{equation}
        p_{ij} = \frac{ \1\{I_i \ne I_j\} \cdot (1 - \vec{\vy}_i^\top \vec{\vy}_j) }
        { \sum_{k=1}^M \1\{I_i \ne I_k\} \cdot (1 - \vec{\vy}_i^\top \vec{\vy}_k) }
        .
    \end{equation}

Given the density function $p_{ij}$, sampling the $N$ pixels without replacement can be efficiently implemented with the Gumbel top-$k$ trick \cite{kool2019stochastic,jang2016categorical}. We omit the detail here.

%% file: sec_exp.tex
\section{Experiment}

\subsection{Datasets and Settings}

\paragraph{VOC 2012.} The PASCAL VOC 2012 segmentation dataset~\cite{everingham2010pascal} comes with a finely labeled \textsc{train} split of 1,464 images, a \textsc{val} split of 1,449 images, and a coarsely labeled \textsc{aug} split of 9,118 images. We use the public 1/2, 1/4, 1/8, and 1/16 subsets of \textsc{train} in \cite{zou2020pseudoseg} as the labeled data. The \textsc{train} and \textsc{aug} sets are combined as the unlabeled dataset. 
We also compare with existing semi-supervised semantic segmentation methods under the 1.5k/9k split setting \cite{souly2017semi,hung2019adversarial,ouali2020semi,zou2020pseudoseg}, which treats \textsc{train} as the labeled set and \textsc{train}+\textsc{aug} as the unlabeled set. VOC contains 20 object categories (excluding the background class). The performance is evaluated by the \emph{mean intersection-over-union} (mIoU) metric on the \textsc{val} set.

\paragraph{Cityscapes.} Cityscapes~\cite{cordts2016cityscapes} contains images of urban driving scenes. The \textsc{train\_fine} split has 2,975 training images, and \textsc{val\_fine} has 5,000 images. We employ the public 1/4, 1/8, and 1/30 subsets of the \textsc{train\_fine} images in recent literature \cite{hung2019adversarial,mittal2019semi,fenga2020dmt,french2020semi,zou2020pseudoseg,olsson2021classmix} as the labeled data, while all images of the original \textsc{train\_fine} are used as the unlabeled data. We report the mIoU over the 18 foreground categories in the Cityscapes \textsc{val\_fine} set.

\paragraph{COCO.} COCO~\cite{lin2014microsoft} is a challenging detection and segmentation dataset of 80 categories. It is significantly larger (118,287 images in the official \textsc{train} set) and more diverse than VOC or Cityscapes. We reuse the same 1/32, 1/64, 1/128, 1/256, and 1/512 splits of the \textsc{train} set from \cite{zou2020pseudoseg} as the labeled data, and further enrich the benchmark by constructing the uniformly downsampled 1/8 and 1/16 labeled variants. We again evaluate the performance by the mIoU metric on the official \textsc{val} set (5,000 images).

\begin{table}[tb]
    \small
    \centering
    \caption{VOC 2012 1/2-1/16 labeled \textsc{train} set results. The settings follow \cite{zou2020pseudoseg}. The numbers of labeled images are shown inside the parentheses. R50 and R101 refer to ResNet 50 and 101. We measure the mIoU on the VOC 2012 \textsc{val} set. The mIoUs of compared methods are the reproduced results in \cite{zou2020pseudoseg} under these data splits. The results of \cite{zou2020pseudoseg,french2020semi} and our method are obtained by DeepLab-v3+; \cite{ouali2020semi} uses PSP-Net whose base performance is comparable to DeepLab-v3+, while other methods use DeepLab-v2. \ours{} obtains favorable results over existing methods. Due to the high variance, we also include the standard deviation of 3 runs on differently sampled 1/16 splits.} 
    \label{tab:voc1}
    \setlength{\tabcolsep}{0.9pt}
    \begin{tabular}{llcccc}
    \toprule
        Method    & Net & 1/2 (732) & 1/4 (366) & 1/8 (183) & 1/16 (92) \\
    \midrule
        AdvSemSeg \cite{hung2019adversarial}  & R101  & 65.27  & 59.97  & 47.58  & 39.69 \\
        CCT \cite{ouali2020semi}  & R50  & 62.10  & 58.80  & 47.60  & 33.10 \\
        MT \cite{tarvainen2017mean}  & R101  & 69.16  & 63.01  & 55.81  & 48.70 \\
        GCT \cite{ke2020guided}  & R101  & 70.67  & 64.71  & 54.98  & 46.04  \\
        VAT \cite{miyato2018virtual}  & R101  & 63.34  & 56.88  & 49.35  & 36.92  \\
        CutMix \cite{french2020semi}  & R101  & 69.84  & 68.36  & 63.20  & 55.58  \\
        PseudoSeg \cite{zou2020pseudoseg} & R50  & 70.42  & 64.85  & 61.88  & 54.89  \\
        PseudoSeg \cite{zou2020pseudoseg} & R101 & 72.41  & 69.14  & 65.50  & \textbf{57.60}  \\
    \midrule
        Sup. baseline  & R50  & 65.73  & 57.76  & 49.57  & 43.97 \\
        Sup. baseline  & R101 & 66.28  & 60.02  & 49.83  & 40.02 \\
        \ours{} (Ours) & R50  & 70.90  & 67.62  & 64.63  & 56.90 $\pm$1.30 \\
        \ours{} (Ours) & R101 & \textbf{73.05}  & \textbf{69.78}  & \textbf{66.28}  & \textbf{57.00 $\pm$1.32} \\
    \bottomrule
    \end{tabular}
\end{table}

\begin{table}[tb]
    \small
    \centering
    \caption{VOC 2012 1.5k/9k split results (\textsc{train} 1.5k images as labeled data and \textsc{train} 1.5k + \textsc{aug} 9k as unlabeled data). The numbers of other methods are directly taken from the corresponding publications.}
    \label{tab:voc2}
    \setlength{\tabcolsep}{10pt}
    \begin{tabular}{llcccc}
    \toprule
        Method    & Network & mIoU (\%)  \\
    \midrule
        GANSeg \cite{souly2017semi} & VGG-16 & 64.10  \\
        AdvSemSeg \cite{hung2019adversarial} & ResNet-101  & 68.40  \\
        CCT \cite{ouali2020semi} & ResNet-50  & 69.40 \\
        PseudoSeg \cite{zou2020pseudoseg} & ResNet-50  & 71.00 \\
        PseudoSeg \cite{zou2020pseudoseg} & ResNet-101 & 73.23 \\
    \midrule
        Sup. baseline  & ResNet-50  & 68.81 \\
        Sup. baseline  & ResNet-101 & 72.00 \\
        \ours{} (Ours) & ResNet-50  & \textbf{72.26}  \\
        \ours{} (Ours) & ResNet-101 & \textbf{74.15}  \\
    \bottomrule
    \end{tabular}
    \vspace{-5pt}
\end{table}

\paragraph{Implementation Details.}
We adopt the DeepLab-v3+ architecture \cite{chen2018encoder} as the test bed, which consists of a fully convolutional backbone, an atrous spatial pyramid pooling (ASPP) based encoder, and a shallow decoder.
We study different backbones including ResNet-50 (the DeepLab modified beta variant \cite{chen2018encoder}), ResNet-101 \cite{he2016deep}, and Xception-65 \cite{chollet2017xception}, to test the robustness of our approach to backbone variations.
The backbones are initialized from their ImageNet \cite{deng2009imagenet} pretrained weights.

The hyper-parameters of our approach mainly include the loss coefficients, the negative sampling strategies, and other design choices of the pixel contrastive loss. They are tuned with the VOC 1/8 split and ResNet-50. Please refer to the ablation study in Sec.~\ref{sec:exp/analysis} for details. These hyper-parameters are used for other benchmarks without further tuning. The performance could be further improved by dataset-specific hyper-parameter selection. In the main results, we set the loss coefficients as $\lambda_1=0.3$ and $\lambda_2=1$. The contrastive loss is computed for each anchor pixel in the last backbone feature maps right before the encoder network (\ie, Conv5 of ResNets and ExitFlow2 of Xception) of both the weak and strong views. The number of negative pixels is 200. They are sampled from both views with the ``Different Image + Pseudo Label'' strategy. Additional training details and DeepLab-related hyper-parameters are presented in the supplementary material.

\begin{table}[tb]
    \small
    \centering
    \caption{Cityscapes experiment results. The labeled data is varied from 1/30 to full of the original \textsc{train\_fine} set. We evaluate the mIoU on the \textsc{val\_fine} set. The result of \cite{french2020semi} is reproduced by \cite{zou2020pseudoseg} based on DeepLab-v3+, while the results of \cite{hung2019adversarial,mittal2019semi,fenga2020dmt,olsson2021classmix} are their reported numbers based on DeepLab-v2. R50 and R101 refer to ResNet 50 and 101. Our approach {\em achieves higher scores than all previous methods}. In the 1/4 and 1/8 settings, even the weaker ResNet-50 backbone performs quite favorably.}
    \label{tab:cityscapes}
    \setlength{\tabcolsep}{1.5pt}
    \begin{tabular}{@{}llcccc}
    \toprule
        Method    & Net & 1 (2,975) & 1/4 (744) & 1/8 (377) & 1/30 (100)  \\
    \midrule
        AdvSemSeg \cite{hung2019adversarial} & R101  & -  & 62.3\phantom{0}  & 58.8\phantom{0}  & -  \\
        s4GAN \cite{mittal2019semi}  & R101  & 65.8\phantom{0}  & 61.9\phantom{0}  & 59.3\phantom{0}  & -  \\
        DMT \cite{fenga2020dmt} & R101  & 68.16  & -  & 63.03  & 54.80  \\
        ClassMix \cite{olsson2021classmix} & R101 & - & 63.63  & 61.35  & -  \\
        CutMix \cite{french2020semi} & R101  & - &  68.33  & 65.82  & 55.71  \\
        PseudoSeg \cite{zou2020pseudoseg} & R101  & - & 72.36  & 69.81  & 60.96  \\
    \midrule
        Sup. baseline  & R50  & 73.64  & 72.86  & 68.06  & 55.25  \\
        Sup. baseline  & R101 & 74.88  & 73.31  & 68.72  & 56.09  \\
        \ours{} (Ours) & R50  & 75.39  & 73.80  & 72.11  & 60.37  \\
        \ours{} (Ours) & R101 & \textbf{75.99}  & \textbf{75.15}  & \textbf{72.29}  & \textbf{62.89}  \\
    \bottomrule
    \end{tabular}
    \vspace{-5pt}
\end{table}

\begin{table*}[tb]
    \small
    \centering
    \caption{COCO results. The original \textsc{train} set is split into 1/512-1/8 subsets to be the labeled data. PseudoSeg \cite{zou2020pseudoseg} results are taken from their paper. \ours{} performs {\em consistently} better than the supervised baseline and \cite{zou2020pseudoseg}. The gain over the supervised baseline is the largest in the 1/256 split, while our 1/8 split result {\em almost matches the supervised full data result}.}
    \label{tab:coco}
    \setlength{\tabcolsep}{2.9pt}
    \begin{tabular}{llcccccccc}
    \toprule
        Method    & Network  & Full (118k) & 1/8 (14786) & 1/16 (7393) & 1/32 (3697) & 1/64 (1849) & 1/128 (925) & 1/256 (463) & 1/512 (232) \\
    \midrule
        Sup. baseline & Xception-65 & 50.10  & 47.91  & 45.12  & 42.24  & 37.80  & 33.60  & 27.96  & 22.94 \\
        PseudoSeg \cite{zou2020pseudoseg}  & Xception-65 & -  & -  & -  & 43.64  & 41.75  & 39.11  & 37.11  & 29.78  \\
        \ours{} & Xception-65 & \textbf{50.67}  & \textbf{49.91}  & \textbf{48.07}  & \textbf{46.05}  & \textbf{43.67}  & \textbf{40.12}  & \textbf{37.53}  & \textbf{29.94}  \\
    \bottomrule
    \end{tabular}
    \vspace{-10pt}
\end{table*}

\subsection{Main Results}

\paragraph{VOC 2012.}
We report the results on the 1/2-1/16 labeled splits in Tab.~\ref{tab:voc1}. Our method (\ours{}) outperforms the prior state-of-the-art (PseudoSeg \cite{zou2020pseudoseg}) and other consistency-based approaches in almost all data splits with the ResNet-50 and ResNet-101 backbones. Note that the ratio of labeled data is small in this setting, since we use the entire \textsc{train}+\textsc{aug} 10,575 images as the unlabeled data. We observe that the gain of our approach is {\em particularly noteworthy for moderately low-data regimes} such as the 1/2 and 1/4 splits. The more extreme case of 1/16 labeled split (92 labeled images) is quite challenging, as there is very little supervision signal for the semi-supervised model to learn in the first place. We thus augment the weak branch with the momentum averaging technique as in MoCo (momentum 0.99) and report the mean and standard deviation (std) of 3 runs of the 3 randomly generated 1/16 splits and observe that our method is {\em comparable} to \cite{zou2020pseudoseg}.

We also compare the results of the 1.4k/9k split in Tab.~\ref{tab:voc2}.
We find that our method outperforms the prior arts under this setting, achieving $74.15\%$ mIoU with ResNet-101.

\paragraph{Cityscapes.}
The Cityscapes results are shown in Tab.~\ref{tab:cityscapes}. Across all semi-superivsed settings, \ours{} outperforms the supervised baseline by a large margin. Notably, with ResNet-101, the mIoU gap between the full set result ($75.99\%$) and the 1/4 labeled setting result ($75.15\%$) is only $0.84\%$. The performance of our method with ResNet-50 is surprisingly good in the 1/8 and 1/30 settings, suggesting that the deeper ResNet-101 backbone might be over-fitting in such settings due to the limited amount of labeled data.

\paragraph{COCO.}
For the COCO experiments, we mainly compare with the recent state-of-the-art PseudoSeg method \cite{zou2020pseudoseg} following their experiment protocols. Xception-65 \cite{chollet2017xception}, a stronger backbone than ResNet-101, is used. The results are shown in Tab.~\ref{tab:coco}. We observe that our method {\em performs better in all data splits}. We even see gains in the full data setting, possibly because of the stronger data augmentation and regularization in our method. The \textsc{val} mIoU of the 1/8 labeled setting is quite promising, {\em almost catching the supervised learning result of the fully labeled data}.

\subsection{Analyses}
\label{sec:exp/analysis}

We report the ablation studies and diagnosis experiments in this section. Without special mentioning, all experiments are conducted with the ResNet-50 backbone and VOC 1/8 labeled split. The mIoUs are evaluated on the VOC \textsc{val} set.

\paragraph{Negative Sampling Strategy.}
One key technical contribution of our method is the negative sampling strategy.
We measure the average False Negative Rates (FNR) during training with different negative sampling strategies in Tab.~\ref{tab:abl/sample}. 
We notice that FNR reduces with more complex sampling distributions. As expected, the Uniform strategy delivers the worst performance, suggesting that the noise introduced by false negative pixels may indeed hurt performance. The ideal case (marked as Oracle) pretends that the algorithm knows the actual ground-truth masks during contrastive learning and can be seen as an upper bound. Our best sampling strategy, Different Image + Pseudo Label, achieves an mIoU {\em almost as good as the Oracle version}.

\begin{table}[tb]
    \small
    \centering
    \caption{Quantitative comparison of negative sampling strategies in VOC 1/8 setting. We report the \textsc{val} mIoU and the False Negative Rates (FNR) of the sampled negative pixels. Oracle refers to the ideal case of using the ground-truth masks to guide the (uniform) negative pixel sampling.}
    \label{tab:abl/sample}
    \vspace{-5pt}
    \setlength{\tabcolsep}{5pt}
    \begin{tabular}{l|llll|l}
    \toprule
        Strategy  & Unif & Diff & Unif+Pseu & Diff+Pseu & Oracle  \\
    \midrule
        mIoU (\%)  & 62.70  & 63.33  & 64.38  & 64.63  & 64.78  \\
        FNR (\%)   & 14.77  & \phantom{0}3.42  & \phantom{0}2.83   & \phantom{0}2.80   & \phantom{0}0 \\
    \bottomrule
    \end{tabular}
    \vspace{-5pt}
\end{table}

\begin{table}[tb]
    \small
    \centering
    \caption{Study on the number of sampled negative pixels. None refers to not using the feature-space pixel contrastive loss. As the number of negatives increases, the \textsc{val} mIoU rises, but the computation and the memory costs also rise. In our experiment, $N=1,600$ actually causes the out-of-memory error. Therefore, we choose $N=200$ in the main results as a trade-off between accuracy and efficiency.}
    \label{tab:abl/numneg}
    \vspace{-5pt}
    \setlength{\tabcolsep}{3.5pt}
    \begin{tabular}{lccccccc}
    \toprule
        $N$ & None  & 50  & 100  & 200  & 400  & 800 & 1600 \\
    \midrule
    mIoU (\%)      & 63.75  & 64.08  & 64.24  & 64.63  & 64.84  &  65.03 & OOM \\
    \bottomrule
    \end{tabular}
    \vspace{-10pt}
\end{table}

\paragraph{Number of Negative Pixels.}
Another importance factor in our pixel contrastive loss is the number of sampled negative pixels $N$. Previous work suggests that a large number of negatives may be beneficial to image-level contrastive learning \cite{chen2020simple,he2020momentum}. While we generally agree with this argument, using a large number of negative pixels can be costly in semantic segmentation. We believe that there should be a trade-off between efficiency and accuracy. In Tab.~\ref{tab:abl/numneg}, we study the effect of the number of negative pixels. With the help of the negative sampling strategy to reduce false negative rates, our approach can already obtain competitive performance with only $N=200$ negative pixels per anchor.

\paragraph{Loss Coefficients.}
We show the impact of the coefficients on the label consistency loss (Eq.~\ref{eq:consist}) and the feature contrastive loss (Eq.~\ref{eq:contrast}) in Tab.~\ref{tab:abl/coef}. We found that $\lambda_1=0.3$ and $\lambda_2=1$ achieve the best result, therefore adopting these values in \emph{all} other data splits and architectures. {\em The coefficients transfer well to other settings}.
Another important observation with this hyper-parameter study is the {\em complementary nature of the label consistency and the feature contrastive properties}. We notice that the joint contrastive-consistent learning ($64.63\%$ when $\lambda_1=0.3, \lambda_2=1$) outperforms either the purely consistent version ($63.75\%$ when $\lambda_1=0, \lambda_2=1$) or the best purely contrastive version ($53.53\%$ when $\lambda_1=0.9, \lambda_2=0$), supporting our insight.

\begin{table}[tb]
    \footnotesize
    \centering
    \caption{Effect of different combinations of unlabeled loss weights. As in Eq.~\ref{eq:unlabeled}, $\lambda_1$ (column) is the coefficient on the pixel contrastive loss in the feature space, $\lambda_2$ (row) is the coefficient on the consistency loss in the label space. Performance is measured by the VOC \textsc{val} mIoU when trained in the 1/8-labeled setting. $\lambda_1=0$ is the special case of applying only label consistency training, while $\lambda_2=0$ is the special case of applying only feature contrastive learning.}
    \label{tab:abl/coef}
    \setlength{\tabcolsep}{6.5pt}
    \begin{tabular}{lcccccc}
    \toprule
      & $\lambda_1=0$ & 0.1 & 0.3 & 0.5 & 0.7 & 0.9  \\
    \midrule
    $\lambda_2=0$ & 49.57 & 51.92 & 52.29 & 52.28 & 52.74 & 53.53 \\
    $\lambda_2=0.5$ & 62.31 & 61.62 & 62.47 & 63.83 & 63.07 & 63.64 \\
    $\lambda_2=0.7$ & 63.46 & 63.26 & 63.71 & 63.79 & 63.35 & 63.43 \\
    $\lambda_2=1.0$ & 63.75 & 63.60 & \textbf{64.63} & \textbf{64.48} & 62.55 & 63.15 \\
    $\lambda_2=1.2$ & 61.33 & 64.48 & 64.30 & 64.16 & 63.08 & 63.03 \\
    \bottomrule
    \end{tabular}
    \vspace{-5pt}
\end{table}

\paragraph{Computational Cost.}
We report the training time comparison in Tab.~\ref{tab:abl/time}. While achieving higher performance, the training time of PC$^2$Seg is comparable to the previous state-of-the-art semi-supervised method PseudoSeg \cite{zou2020pseudoseg}. If we remove the contrastive component of PC$^2$Seg, the time is reduced by less than 5min, suggesting that the extra cost of our pixel contrastive learning is low.
We further show the cost reduced by our negative sampling strategy through a simple calculation in the supplementary material.

\begin{table}[tb]
\centering
\footnotesize
\caption{Training time comparison in the VOC 1/8 setting.}
\label{tab:abl/time}
\vspace{-5pt}
\setlength{\tabcolsep}{4.5pt}
\begin{tabular}{lcccc}
\toprule
    & {\scriptsize Supervised} & {\scriptsize PseudoSeg} & {\scriptsize PC$^2$Seg consist-only} & {\scriptsize PC$^2$Seg}
       \\
\midrule
    Time (min)
    & 38  & 80  & 75$\sim$80  &  80  \\
    \textsc{val} mIoU (\%)
    & 49.57  & 61.88  & 63.21  &  64.63  \\
\bottomrule
\end{tabular}
\vspace{-10pt}
\end{table}

\paragraph{Comparison to Other Label-Space and Feature-Space Losses.}
In the label space, we compare two variants: the normalized \ltwo loss (Sec.~\ref{sec:method}) and the cross-entropy (CE) loss.
In the feature space, we compare four variants: (1) no feature-space loss (None), (2) image contrastive loss \cite{chen2020simple}, (3) pixel consistency loss, which is the normalized \ltwo distance between pixel features, and (4) pixel contrastive loss. Tab.~\ref{tab:abl/losses} shows that the combination of the label \ltwo loss and the feature pixel contrastive loss achieves the best result.

\begin{table}[tb]
    \small
    \centering
    \caption{Investigation of the variants that use different loss functions on the label space (row) and the feature space (column) with other hyper-parameters fixed. Overall, we observe that the \ltwo loss is more robust as a label-space consistency loss than the cross-entropy (CE) loss. In the 2nd row, the pixel contrastive loss outperforms the pixel consistency loss and the image-level contrastive loss.}
    \label{tab:abl/losses}
    \vspace{-5pt}
    \setlength{\tabcolsep}{4.5pt}
    \begin{tabular}{lcccc}
    \toprule
    mIoU (\%) &  None &  Img Contrast  & Pix Consist  &  Pix Contrast  \\
    \midrule
    Output CE  & 63.54  & 60.41  & 59.33  & 58.47  \\
    Output \ltwo   & 63.75  & 62.49  & 63.21  & \textbf{64.63}  \\
    \bottomrule
    \end{tabular}
    \vspace{-5pt}
\end{table}

\paragraph{Choice of Contrastive Learning Layer.}
In the main results, we choose to apply pixel contrastive learning on the last feature map of the backbone networks (Conv5 block of ResNet and ExitFlow2 block of Xception). We show results of applying it on other ResNet layers in Tab.~\ref{tab:abl/layer}. Earlier stage feature maps are larger in size and contain more low-level details, while later stage feature maps lose resolutions but contain more semantics. A mid-level intermediate layer yields the best performance.

\begin{table}[tb]
    \small
    \centering
    \caption{Varying the feature layer to apply the pixel contrastive loss. We report the VOC \textsc{val} mIoU in the 1/8 setting. Conv3-5 are the corresponding conv blocks of ResNet-50. Encoder refers to the feature map produced by the DeepLab ASPP net. Decoder refers to the final feature map right before classification. Conv5+Enc imposes a contrastive loss on two best-performing layers (Conv5 and Encoder), which does not bring extra gains over a single layer.}
    \label{tab:abl/layer}
    \vspace{-5pt}
    \setlength{\tabcolsep}{1.9pt}
    \begin{tabular}{lcccccc}
    \toprule
        Layer  & Conv3 & Conv4 & Conv5 & Encoder & Decoder & Conv5+Enc \\
    \midrule
    mIoU (\%)  &  60.70 & 63.90 & \textbf{64.63} & 64.25  & 63.74  & 64.59  \\
    \bottomrule
    \end{tabular}
    \vspace{-5pt}
\end{table}

\paragraph{Projection Layer.}
The projection layer performs dimensionality reduction on the feature vectors before contrasive learning. We find that using separate projections for the weak and strong branches yields higher mIoU than using a shared projection head ($64.63\%$ \emph{vs.} $63.97\%$).

\paragraph{Delayed-Start of Semi-Supervised Learning.}
We mimic the two-stage variant that delays the start of semi-supervised learning until certain steps in Tab.~\ref{tab:abl/delay}. We notice that breaking the training into two stages in fact decreases the performance. Applying all losses jointly throughout the training ($\text{Delay}=0$) achieves a better result in our experiments.

\begin{table}[tb]
    \small
    \centering
    \caption{Delaying the unlabeled loss until certain steps. The total number of steps is 30,000. Applying all losses jointly without delay achieves the best result.}
    \label{tab:abl/delay}
    \setlength{\tabcolsep}{6.5pt}
    \begin{tabular}{lccccc}
    \toprule
    Delayed Steps  & 0  & 2,000  & 4,000 & 8,000 & 12,000 \\
    \midrule
    mIoU (\%)   & \textbf{64.63}  & 63.30 & 62.67 & 62.56 & 62.50 \\
    \bottomrule
    \end{tabular}
    \vspace{-10pt}
\end{table}

%% file: sec_supp.tex
\appendix

\numberwithin{figure}{section}
\numberwithin{table}{section}

\section{Additional Implementation Details}

In Tab.~\ref{tab:hyperparam}, we list the DeepLab-related hyper-parameters. They take either the recommended default values by \cite{chen2018encoder} or the values from the public code of \cite{zou2020pseudoseg}. The learning rate is linearly annealed from $0.007$ to $0$ as training proceeds. The weight decay is 1e-4 for ResNets \cite{he2016deep} and 4e-5 for Xception \cite{chollet2017xception}. The input train crop size is set to $513\times 513$ for all datasets. The eval crop size is set to $513\times 513$ for VOC \cite{everingham2010pascal}, $641\times 641$ for COCO \cite{lin2014microsoft}, and $1,025\times 2,049$ for Cityscapes \cite{cordts2016cityscapes}. The DeepLab encoder output stride is chosen as $16$. In the VOC experiments, the model is trained for 30,000 gradient steps that are distributed on 4 asynchronous replica workers. Each worker has 2 GPUs. The Cityscapes and COCO experiments require longer training duration, which takes a total of 130,000 and 200,000 steps, respectively, on 8 asynchronous workers of 2 GPUs. For both the labeled data and unlabeled data branches, the training batch size per GPU is 4, or equivalently 8 images per worker.

\begin{table}[h]
    \small
    \centering
    \caption{Summary of hyper-parameter settings. Hyper-parameters take either the recommended default values by \cite{chen2018encoder} or the values from the public code of \cite{zou2020pseudoseg}.}
    \label{tab:hyperparam}
    \setlength{\tabcolsep}{2.5pt}
    \begin{tabular}{@{\,}llll}
    \toprule
        Hyper-parameter & VOC  & Cityscapes  & COCO  \\
    \midrule
        Network         & ResNet50/101 & ResNet50/101 & Xception65  \\
        Weight decay    & 1e-4  & 1e-4  & 4e-5  \\
        Train crop size & $513 \times 513$ & $513 \times 513$ & $513 \times 513$  \\
        Eval crop size  & $513 \times 513$  & $1,025 \times 2,049$  & $641 \times 641$  \\
        Output stride   & 16  & 16  & 16  \\
        Atrous rates    & [6, 12, 18]  & [6, 12, 18]  & [6, 12, 18]  \\
        Train steps     & 30,000  & 130,000  & 200,000 \\
        Learning rate   & $0.07 \to 0$  & $0.07 \to 0$  & $0.07 \to 0$  \\
        Dist. workers   & 4 &  8  & 8  \\
        GPUs/worker     & 2 V100(16G)  & 2 V100(16G)  & 2 V100(16G)  \\
        Batch size/GPU  & 4  & 4  & 4  \\
    \bottomrule
    \end{tabular}
\end{table}

\section{Additional Ablation Results}

\begin{table}[tb]
    \small
    \centering
    \caption{Ablation study on the dimension of the feature projection layer in the VOC 1/8 split and ResNet-50 setting.}
    \label{tab:abl/proj_dim}
    \vspace{-5pt}
    \setlength{\tabcolsep}{10pt}
    \begin{tabular}{lcccc}
    \toprule
        Dimension    & 64  & 128  & 256 &  512  \\
    \midrule
        \textsc{val} mIoU (\%)    & 64.12  & \textbf{64.63}  & 64.07 &  63.74 \\
    \bottomrule
    \end{tabular}
\end{table}

\vspace{-7pt}
\paragraph{Dimension of Projection Layer.} The impact of the dimension of the projection layer in our \ours{} method is studied in Tab.~\ref{tab:abl/proj_dim} under the VOC 1/8 ResNet-50 setting. For the results reported in the main paper, we chose the dimension as 128. We found that alternative values do not bring gains over the chosen value if other hyper-parameters are fixed. Alternative dimension values may require re-tuning some hyper-parameters to achieve better results.

\begin{table}[tb]
    \small
    \centering
    \caption{Comparison between the supervised baseline, the label-consistent-only version (denoted as `\ours{} w/ LC') and the joint label-consistent and feature-contrastive version of \ours{} with the VOC splits and ResNet-50 backbone.}
    \label{tab:abl/voc_consist}
    \vspace{-5pt}
    \setlength{\tabcolsep}{1.5pt}
    \resizebox{.48\textwidth}{!}{
    \begin{tabular}{@{}lccccc}
    \toprule
        Method/Split  & 1 (1464)  & 1/2 (732) & 1/4 (366) & 1/8 (183) & 1/16 (92) \\
    \midrule
        Supervised &  68.81  & 65.73  & 57.76  & 49.57  & 43.97  \\
        \ours{} w/ LC (Ours)  & 71.95  & 70.88  & 66.71  & 63.75  & 56.32  \\
        \ours{} (Ours) & \textbf{72.26}  & \textbf{70.90}  & \textbf{67.62}  & \textbf{64.63}  & \textbf{56.90}  \\
    \bottomrule
    \end{tabular}}
\end{table}

\begin{table}[tb]
    \small
    \centering
    \caption{Comparison between the supervised baseline, the label-consistent-only version (denoted as `\ours{} w/ LC') and the joint label-consistent and feature-contrastive version of \ours{} with the Cityscapes 1/8 split and ResNet-50 backbone.}
    \label{tab:abl/city_consist}
    \vspace{-5pt}
    \setlength{\tabcolsep}{1.9pt}
        \resizebox{.48\textwidth}{!}{
    \begin{tabular}{lccc}
    \toprule
        Method  & Supervised  & \ours{} w/ LC (Ours) & \ours{} (Ours)  \\
    \midrule
        \textsc{val\_fine} mIoU (\%) & 68.06  & 71.79  & 72.11  \\
    \bottomrule
    \end{tabular}}
    \vspace{-5pt}
\end{table}

\paragraph{Label Consistent Only.} In Tab.~9 and Paragraph ``Comparison to Other Label-Space and Feature-Space Losses" of Sec.~4.3 in the main paper, we have compared \ours{} with its label-consistent-only version in the VOC 1/8 split setting. Here, we provide additional results with other data splits, which support the claim that the joint label-consistent and feature-contrastive regularization performs better than the label-consistent regularization alone. The label-consistent version essentially removes the pixel contrastive loss and keeps everything else unchanged. The VOC results are shown in Tab.~\ref{tab:abl/voc_consist}. We have a similar observation with the Cityscapes 1/8 split in Tab.~\ref{tab:abl/city_consist}, where the label-consistent-only version achieves 71.79\% validation mIoU in comparison to 72.11\% mIoU of full \ours{} with the joint contrastive-consistent regularization.

\zyy{
\section{Training Time and Computational Cost}
}

Since there is an additional unlabeled data branch in our semi-supervised method, the training inevitably takes longer time than the purely supervised baseline. As stated in the main paper, we measure the training time in the VOC ResNet-50 experiments. Our implementation of the supervised baseline took 38 minutes, while our \ours{} with label consistency and feature contrastive learning took roughly 80 minutes, and the label-consistent-only version of \ours{} took around 75 to 80 minutes. Such training time is comparable to existing approaches {\em within the semi-supervised setting} -- \eg, state-of-the-art PseudoSeg \cite{zou2020pseudoseg} also took about 80 minutes.

We further show the cost reduced by our negative sampling strategy through a simple calculation. The feature tensor shape in practice is $[4,33,33,128]$ for a batch of 4 images (512-by-512 pixels). If using all pixels as negatives, we need to compute a $[4\cdot33^2, 4\cdot33^2]$ inner-product matrix ($19M$ elements) with $(4\cdot33^2)^2 \cdot 128 = 2,428M$ MulAdd operations. But, if we only draw $200$ negative pixels, it is reduced to computing a $[4\cdot33^2, 200]$ matrix ($0.8M$ elements) with $4\cdot33^2 \cdot 200 \cdot 128 = 111M$ operations. Considering the negative sampling itself requires $(4\cdot33^2)^2 \cdot 20=379M$ operations to compare the pseudo labels (20 is the number of VOC classes), the overall floating point operations are about 5 times fewer.

\section{Visualization}

\begin{figure}
    \centering
    \includegraphics[width=\linewidth]{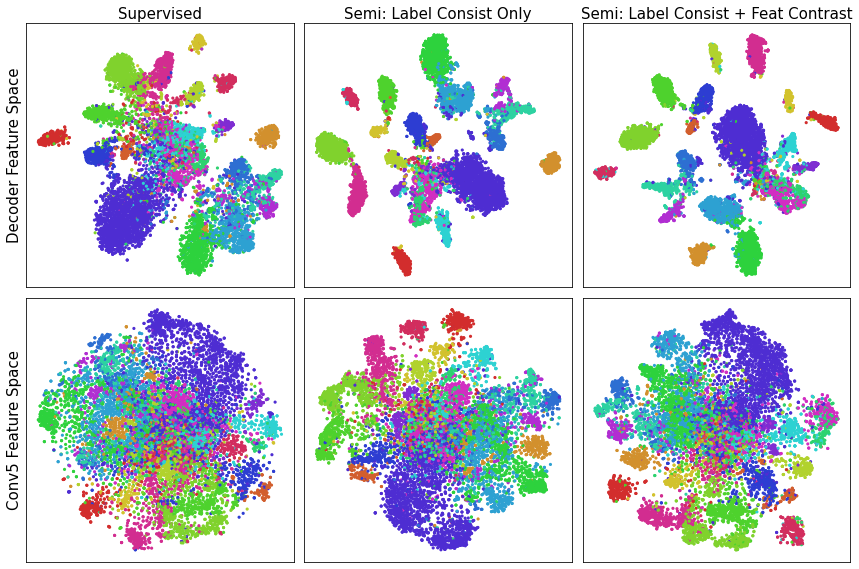}
    \caption{t-SNE \cite{van2008visualizing} visualization of the Decoder and Conv5 (of ResNet-50) feature spaces generated by different methods on the VOC \textsc{val} images. The train set is the VOC 1/8 split. Conv5 is the feature layer where the pixel contrastive loss is applied. We randomly sample 10,000 data points to produce the t-SNE plot, and the perplexity parameter is set to 40. We observe clearer separation of semantic classes in the feature spaces with semi-supervised methods than that with the supervised baseline.}
    \label{fig:voc_tsne}
    \vspace{-5pt}
\end{figure}

\begin{figure*}[p]
    \centering
    \includegraphics[width=\linewidth]{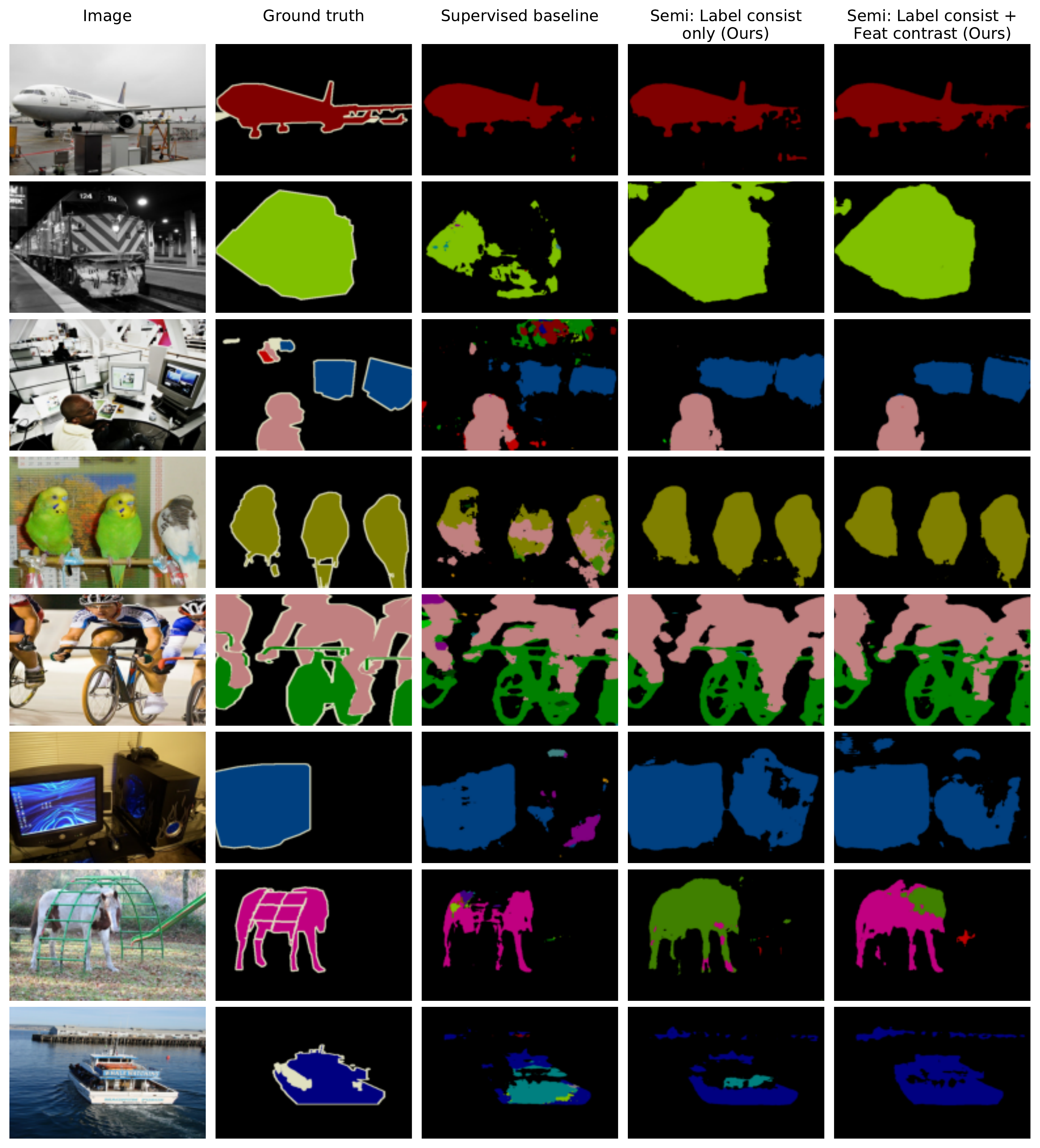}
    \caption{VOC 2012 prediction results. Models are trained in the 1/8 split setting. Images are from the \textsc{val} set. The white pixels in the ground-truth masks indicate ignored regions. {\em Overall, both the label-consistent-only and the contrastive-consistent variants of our semi-supervised method produce significantly less noisy predictions than the supervised baseline}. The consistency regularization is able to suppress some background artifacts as in the 3rd row example. Some other success cases include the airplane in the 1st row and the train in the 2nd row, where our final method covers fuller extents of the objects. Some failure cases include the monitor in the 3rd to last row, where the model mistakenly classifies the (possibly co-occurring) machine into the monitor class, the mis-classification of the horse in the 2nd to last row, and the false positives in the last row.}
    \label{fig:voc_pred}
\end{figure*}

\begin{figure*}[p]
    \centering
    \includegraphics[width=\linewidth]{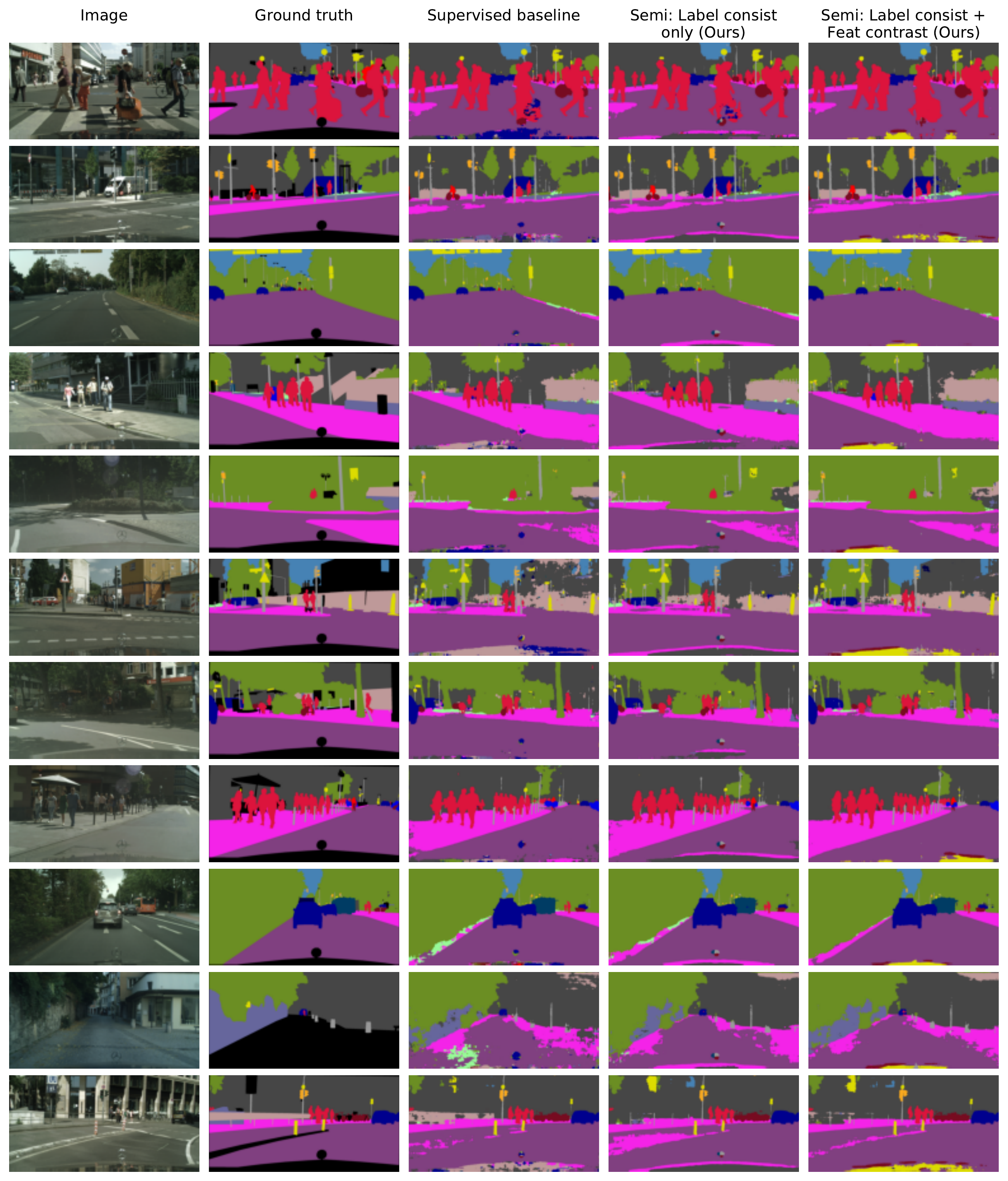}
    \caption{Cityscapes prediction results. Models are trained in the 1/8 split setting. Images are from the \textsc{val\_fine} set. The dark areas in the ground-truth masks indicate ignored regions. We notice that there exist some {\em substantially higher-quality} cases of our contrastive-consistent learning based semi-supervised method over others, such as the person's bag in the 1st row, the sidewalk in the 2nd row, and the upper traffic sign in the 3rd row.}
    \label{fig:city_pred}
\end{figure*}

To have a better understanding of \ours{}, we use two complementary approaches to visualize the results: (1) the t-SNE plot \cite{van2008visualizing} of the feature spaces in Fig.~\ref{fig:voc_tsne}, and (2) the predicted segmentation masks in Fig.~\ref{fig:voc_pred} and Fig.~\ref{fig:city_pred}.

From the t-SNE plot, we observe that both the label-consistent-only and the joint contrastive-consistent variants of our method generate feature spaces that are more separable than the supervised baseline. In the decoder feature space, joint contrastive-consistent variant seems to increase slightly the margins between a few categories, compared with the label-consistent-only variant. From the predicted masks, we can see some success and failure cases of \ours{}. Please refer to the figure captions for detailed descriptions.

